# Building for Tomorrow: Assessing the Temporal Persistence of Text Classifiers

Rabab Alkhalifa[a,b,*], Elena Kochkina[a,c] and Arkaitz Zubiaga[a]

[a]School of Electronic Engineering and Computer Science, Queen Mary University of London, Mile End Road, London, E1 4NS, United Kingdom
[b]College of Computer Science and Information Technology, Imam Abdulrahman Bin Faisal University., Dammam, P.O.Box 1982, Eastern Province, Kingdom of Saudi Arabia
[c]Alan Turing Institute, 2QR, John Dodson House, 96 Euston Rd, London, NW1 2DB, United Kingdom

ABSTRACT

Performance of text classification models tends to drop over time due to changes in data, which limits the lifetime of a pretrained model. Therefore an ability to predict a model's ability to persist over time can help design models that can be effectively used over a longer period of time. In this paper, we provide a thorough discussion into the problem, establish an evaluation setup for the task. We look at this problem from a practical perspective by assessing the ability of a wide range of language models and classification algorithms to persist over time, as well as how dataset characteristics can help predict the temporal stability of different models. We perform longitudinal classification experiments on three datasets spanning between 6 and 19 years, and involving diverse tasks and types of data. By splitting the longitudinal datasets into years, we perform a comprehensive set of experiments by training and testing across data that are different numbers of years apart from each other, both in the past and in the future. This enables a gradual investigation into the impact of the temporal gap between training and test sets on the classification performance, as well as measuring the extent of the persistence over time. Through experimenting with a range of language models and algorithms, we observe a consistent trend of performance drop over time, which however differs significantly across datasets; indeed, datasets whose domain is more closed and language is more stable, such as with book reviews, exhibit a less pronounced performance drop than open-domain social media datasets where language varies significantly more. We find that one can estimate how a model will retain its performance over time based on (i) how well the model performs over a restricted time period and its extrapolation to a longer time period, and (ii) the linguistic characteristics of the dataset, such as the familiarity score between subsets from different years. Findings from these experiments have important implications for the design of text classification models with the aim of preserving performance over time.

## 1. Introduction

A supervised text classification model relies on labelled datasets to train the model (Sebastiani, 2002). From an experimental perspective, the design and evaluation of classification models typically rely on data pertaining to fixed periods of time. Recent research demonstrates that such models, while showing competitive performance in their experimental environment, underperform when they need to classify new data that is distant in time from that observed during training (Alkhalifa and Zubiaga, 2022). This deterioration of performance has been demonstrated for different classification tasks, including topic classification (Rocha, Mourão, Pereira, Gonçalves, and Meira, 2008), sentiment classification (Lukes and Søgaard, 2018), hate speech detection (Florio, Basile, Polignano, Basile, and Patti, 2020), stance detection (Alkhalifa, Kochkina, and Zubiaga, 2021) and political ideology detection (Röttger and Pierrehumbert, 2021). This performance drop can happen for multiple reasons, including among others the evolution in language use (Smith, 2004) or the evolution of public opinion (Bonilla and Mo, 2019) and its extent may vary (Alkhalifa et al., 2021). This poses an important challenge and limitation on such models when one plans to continue using the model over a long period of time to classify new, incoming data, as can be the case with a stream of user-generated contents (Cheng, Chen, Lee, and Li, 2021).

*Corresponding author
raalkhalifa@iau.edu.sa, r.a.a.alkhalifa@qmul.ac.uk (R. Alkhalifa); a.zubiaga@qmul.ac.uk (A. Zubiaga)
http://www.zubiaga.org/ (A. Zubiaga)
ORCID(s): 0000-0002-2875-5400 (R. Alkhalifa); 0000-0003-0691-3647 (E. Kochkina); 0000-0003-4583-3623 (A. Zubiaga)

Despite this evidence of performance deterioration over time, previous research has not explored the nature of this deterioration, i.e. when, how and why it occurs, such that it could inform design and maintenance of text classification models which can continue to be used as effectively as possible over time. Our study fills this gap by performing a comprehensive study into model performance over time, i.e. by keeping all variables in the experiments fixed, where the only factor that changes is time, with the associated evolution of data over this time. This helps us shed light into the causes of performance deterioration, as well as to devise possible solutions to mitigate this deterioration.

In this paper, by using three large-scale, longitudinal text classification datasets involving user-generated content, we perform a set of experiments to learn more about the temporal persistence of text classifiers. Our study focuses on quantifying the impact of these different factors (representation models, classification algorithms, datasets) as well as helping understand its implications for the design of temporally-persistent text classification models. Our work is the first to delve into the problem of temporal persistence of text classification models.

Through this study, we tackle the research aims defined in Section 1.1 by focusing on the research questions in Section 1.2, and make the novel contributions discussed in Section 1.3.

## 1.1. Research aims

Our study has two overarching aims: (1) assessing how different factors of the datasets and models affect performance over time, and (2) where one only has access to annotated data covering a small timeframe, determining whether one can predict the temporal performance that different classifiers will exhibit over time. An ability to design classifiers after only seeing a short timeframe of annotated data can be very important to enable design of temporally robust classifiers where annotation of longitudinal data is costly and unaffordable.

To address these two larger aims, we break down our research into six smaller research objectives:

1. assessing the temporal persistence of existing **language models**, quantifying their performance drop in different situations,
2. investigating the impact of **classification model** choice on temporal performance,
3. understanding **when and why model performance drops** over time, which informs when a model needs adapting, based on factors including content representation, classification model and dataset characteristics,
4. understanding the **potential and limitations of contextual language models** to develop temporal persistence, which in turn informs annotation practices,
5. looking at **dataset patterns from a longitudinal angle**, to assess the impact of different factors, and
6. assessing how different metrics extracted from small timeframes of the dataset can determine performance over time, such that we can improve **predictability of temporal performance** when longitudinally annotated datasets are not available and/or affordable.

## 1.2. Research questions

Our overarching research question is "how do different factors in the experiment design determine the temporal decay (or lack thereof) of classification results?". We break down this research question into five smaller ones that we address in a set of experiments:

- **RQ1.** How does the choice of a **language model** impact classification performance over time across different datasets?
- **RQ2.** How does the choice of an **algorithmic architecture** impact classification performance over time across different datasets?
- **RQ3.** How does prediction **performance vary across longitudinal datasets of different types**?
- **RQ4.** What are the **linguistic features that, in the absence of sufficient labelled data for direct testing, can help us estimate the temporal persistence on a particular dataset**?
- **RQ5.** How **stable are contextualised language model representations** with respect to temporally **evolving context changes**?

### 1.3. Contributions

In this paper, we make the following novel contributions:

- We perform comprehensive experiments on three longitudinal text classification datasets, with the aim of assessing the factors that impact model performance. We focus on key factors including language representation approaches, classification algorithms and dataset characteristics; concluding how, when and the extent to which they impact classification performance.
- We perform a comprehensive analytical study of dataset characteristics to investigate how factors such as word frequencies in datasets can help predict model performance where one lacks sufficient labelled datasets over time.
- We propose a novel scalable methodology to quantify contextual meaning drift for dynamic (diachronic) aspects over time, which can help assessing how well a contextual model recognises texts from different periods.
- Our study provides insights into the factors that impact model performance drop over time, looking at five key dimensions: language representations, classification algorithms, time, lexical features and context.

We devise a set of best practices to consider when designing text classification models with the aim of keeping their performance as stable as possible over time. Among others, our study highlights that (i) the classifier that shows top performance on test data pertaining to the same time period is likely to consistently perform best over time, whose performance drop is comparable to other models and does not impact the ranking among models, (ii) linguistic variability of the dataset at hand can help make an informed decision on the optimal classifier design, which depending on the amount of data available can be based on quantitative metrics or qualitative estimates (e.g. whether it is a social media dataset and whether it is a quickly evolving domain), and (iii) contextual language models provide a solid methodology to maximise temporal persistence thanks to their capacity to model sub-words, but also show that there is still room for improvement.

### 1.4. Paper structure

This paper is organised as follows. Section 2 introduces background on the problem of temporal performance on NLP classification tasks. In Section 3, we describe our research methodology, including our evaluation settings. We then describe and analyse the longitudinal datasets we use in Section 4, followed in Section 5 by the language models and classification algorithms we used, and lexical analysis methods used. In Section 6, we present our experimental objectives, evaluation and analysis of results. We provide a critical analysis of our findings in Section 7, concluding the paper in Section 8.

## 2. Related work

### 2.1. Overview of research on temporal changes in data

Related areas of research addressing temporal aspects of model performance have focused on evolutionary learning (He, Li, Song, He, and Peng, 2018; Pustokhina, Pustokhin, Aswathy, Jayasankar, Jeyalakshmi, Díaz, and Shankar, 2021) and lifelong machine learning (LLML) (Nguyen, Pham, Nguyen, Nguyen, and Ha, 2020; Ha, Pham, Nguyen, Nguyen, Vuong, Tran, and Nguyen, 2018; Xu, Pan, and Xia, 2020). These works however assume that longitudinally annotated data are available, which can be incorporated into the model for continuous training. This line of research can be complementary to ours, however it is not applicable in our scenario where the data available for training is temporally restricted, i.e. the model needs to make the most of labelled data for a particular period of time which will then be applied to test data distant in time.

Alkhalifa and Zubiaga (2022) provided a theoretical perspective into the impact of time on classification models, delving into a number of challenges that text classification models face when applied to data pertaining to time periods that differ from that used for training. Focusing on the stance detection task in the context of social media, through an analysis of the literature, they attributed the temporal impact on stance detection models to three main factors: stance utterance, stance context and stance influence.

Other works have studied temporal aspects of model performance through empirical experiments. The link between words and classes evolves over time, as a result of terms emerging, disappearing, and exhibiting varying dimensions, e.g. word polarity drift demonstrated by Rocha et al. (2008). Lukes and Søgaard (2018) investigated polarity rank drift over time using a logistic regression classifier. They illustrated how the polarity of words can impact model performance. Another line of research that focuses on concept drift looks into the impact of changes in label distribution over time (Nishida, Hoshide, and Fujimura, 2012), such as hashtag usage change.

| Ref. | Task | Temporal Granularity | Dataset (time-span) |
|---|---|---|---|
| Rocha et al. (2008) | Document classification | Yearly | ACM-DL (1980-2001) and MedLine (1980-2001) |
| Nishida et al. (2012) | Document classification | Daily and hourly | Hashtags as different topics |
| Preoţiuc-Pietro and Cohn (2013) | Hashtag classification | Monthly | Twitter public Gardenhose stream (2011) |
| He et al. (2018) | Topic classification | Monthly and yearly | NYTimes (1996 to 1997), RCV1 (Jan 01, 1987 to June 19, 2007) |
| Lukes and Søgaard (2018) | Review raing prediction | Across group of years | ARR (2001 to 2014) |
| Florio et al. (2020) | Hate speech detection | Monthly | unbalanced Haspeede (Oct 2016 to 25 April 2017), TWITA (2012-2017) |
| Murayama, Wakamiya, and Aramaki (2021) | Fake news detection | Across group of years | MultiFC, Horne17, Celebrity, Constraint (2007–2015, 2016, 2016-2017, 2020) |
| Allein, Augenstein, and Moens (2021) | Fact-checking | Daily | MultiFC |
| Röttger and Pierrehumbert (2021) | Document classification | Monthly | Reddit Time Corpus (RTC) (March 2017 and February 2020), Political Subreddit Prediction (PSP) |
| Alkhalifa et al. (2021) | Stance detection | Yearly | GESD (2014-2019) |
| Lazaridou, Kuncoro, Gribovskaya, Agrawal, Liska, Terzi, Gimenez, de Masson d'Autume, Kocisky, Ruder et al. (2021) | Question answering | Yearly | WMT (2007-2019), ARXIV (1986-2019) and CUSTOMNEWS (1969-2019) |
| **Our work** | Stance detection, sentiment analysis and review raing prediction | Yearly | GESD (2014-2019), TESA (2013-2020) and ABRR (2000-2018) |

**Table 1**
Summary of related works addressing data evolution and model performance over time.

Related to this paper, Alkhalifa et al. (2021) and Röttger and Pierrehumbert (2021) performed preliminary experiments leading to findings into the impact of time on model performance and dataset usability on upstream and downstream tasks. The present study provides a much more comprehensive exploration of the problem by studying three main factors: (1) language representation approaches, (2) classification algorithms and (3) dataset characteristics. We study the impact of time using three datasets spanning between 6 and 19 years.

## 2.2. Empirical research into temporal classification

There is a body of research looking into the impact of time on classification performance, which we summarise in Table 1. These works usually focus only on a single classification task (such as hate speech detection (Florio et al., 2020)), and are mostly limited to analysing only one aspect of the problem, e.g. either datasets or models. Our comprehensive focus on multiple tasks and aspects enables to draw more generalisable insights into the problem with a view on informing future research.

As can be seen in the table, the majority of NLP benchmarks today still do not study temporal generalisability across multiple benchmarks to assess and delve into the impact on different tasks. Previous work investigated **supervised temporal adaptive approaches** that study the impact of annotated dataset size on performance persistence over time (Rocha et al., 2008; Nishida et al., 2012; He et al., 2018; Lukes and Søgaard, 2018; Florio et al., 2020) assuming that annotated data is available for both training and testing time periods. Moreover, Florio et al. (2020) demonstrated that when the **amount of annotated data is progressively augmented** from the target year, some models tend to be more sensitive than others. This differs from our main objective, where we consider scenarios where the annotated data for model training only covers a limited time period. Using temporal gaps between training and test data, we restrict our model's ability to acquire additional annotated data. Our research is unique in that we aim to measure model persistence by assessing by design models' temporal generalisability across a sizable number of dynamically evolving training and test sets.

Several existing works also introduced **unsupervised temporal adaptive approaches** to improve the temporal persistence of text classifiers, either through **incremental static embedding training** (Alkhalifa et al., 2021; He et al., 2018) or through **continuous pretraining** using transformers (Röttger and Pierrehumbert, 2021; Lazaridou et al., 2021). Their objective however differs from ours, as we focus instead on drawing an in-depth understanding of the impact of different aspects on temporal performance by employing widely-used methods. Our objective is to perform a hitherto lacking investigation into the extent to which a model's performance drops due to temporal language evolution Alkhalifa et al. (2021), as well as when and why it occurs, with the aim of devising best practices for the development of models with their temporal persistence as the objective.

This work differs from the above body of work in several ways. *First*, with special focus on three downstream datasets assessment, we explore **a wide range of language models, classical and deep learning classifiers, using balanced and unbalanced datasets**. We argue that it is critical to provide a broad perspective and reveal the strengths, weaknesses and qualities of existing state-of-the-art models when applied on longitudinal datasets. *Second*, we look at **how a text classifier performs in the past and the future** using data splits that are longitudinally distant in time. Furthermore, we **compare performance trends to the initial performance of the testing year**, which was constructed from the same year as training. In our study, we refer to this as "Temporal Gap 0" since there is no time gap between the test and training years. This is based on a typical NLP benchmark dataset that was collected and annotated in the same time frame for both training and test Alkhalifa and Zubiaga (2022). *Third*, we **introduce and explore the use of novel methods for lexical analysis designed for assessing unlabeled longitudinal corpora**. This allows for a labour-unintensive approach of measuring the complexity of the testing set, a fair assessment of the model's temporal generalisability, as well as an assessment of the potential of a model to generalise over time before we see labelled data in the future.

## 3. Methodology

In this section, we describe how we formulate our experiments to enable temporal analysis of classifiers, and define the evaluation metrics we use.

### 3.1. Problem formulation and assumptions

We define temporal persistence as the ability of a classification model to maintain performance levels (to generalise) when evaluated on a testing set distant in time from the data used to train the model. We simulate the scenario where one only has access to labelled training data pertaining to a specific point in time, i.e. data collected in a specific year. Test data can belong to a different period of time (i.e. a different year), which can be more or less distant in time from the data used for training. This simulates the realistic scenario where one has labelled data collected in a specific point in time but the model learned from that data needs to be applied to data in later time periods.

One important aspect when setting up the temporal experiments across years is to control for all confounding factors affecting the datasets (such as topics covered or differences in annotation instructions), so that the only impacting factor is the evolution of time (Rocha et al., 2008; Alkhalifa et al., 2021). We control for this by using datasets collected and labelled uniformly over long periods of time (Section 4). In addition, we re-balance the distributions of labels for consistency across years, i.e. to avoid different label distributions having an impact on performance.

We split the datasets into subsets covering 1-year time intervals. We chose 1-year intervals because the longitudinal nature of our datasets allows us to do so, and because it enables exploration of more substantial changes than if we focused on shorter intervals such as weeks or months. Testing on larger time intervals could be desirable, however we are also limited by the available datasets.

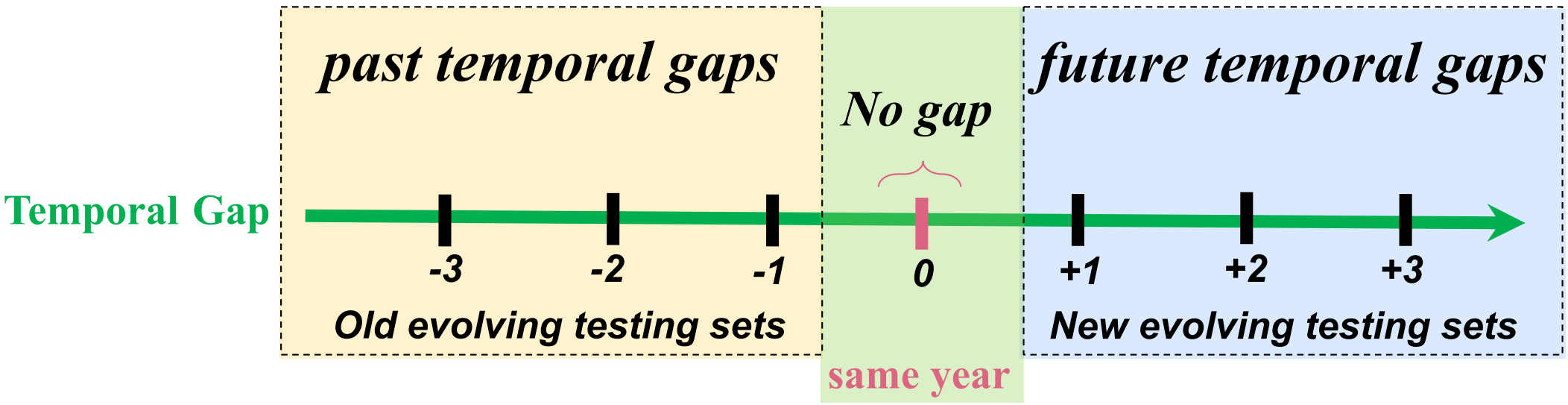

**Figure 1:** An overview of our evolving benchmarks settings used for monitoring temporal generalisability of models' performance.

### 3.2. Evaluating temporal persistence of classification models

First, we define the concept of **Temporal Gap**, which refers to the temporal distance (in years) between training and test data. The temporal gap considers the temporal direction, i.e. negative when the test data is older than the training (past) and positive when the test data is newer than the training (future). Figure 1 illustrates our setup, where **gap 0** (or no gap) represents a testing set pertaining to the same temporal period as the training; **gap -2** represents a testing set that is two years older than the training set, and **gap +2** refers to the testing set from two years after the training set. We consider temporal model persistence for different NLP classification tasks in three settings: (1) learning from older training sets, (2) learning from same-period training sets and (3) learning from newer training sets.

More formally, the temporal gap $G$ is a polarised score which describes the time difference between target year $j$ used for testing and the source year $i$ used for training. The temporal gap is computed for all possible pairs of years within the complete time span in the dataset including $N$ years. The temporal gap ($G$) is defined as follows.

$$G = j - i, \forall i, j \in N \tag{1}$$

where the resulting $G \in [-N + 1, N - 1]$.

In our experiments we first define a source year $i$ and a target year $j$. We then train each model using the training data from year $i$ ($D^i_{tr}$) and test it on the target year $j$ ($D^j_{ts}$). We then iterate over all possible source $i$ and target $j$ year combinations available in the dataset consisting of $N$ years. Finally, we group the results of the experiments by the temporal gap $G$ between years $i$ and $j$ and we present the average of the performance metrics in each of the groups (for each of the time gaps).

For example, if a dataset contains data for the years 2015-2018, temporal gaps would range from -3 to 3, with the following combinations (training year first, followed by test year):

- **G = -3:** 2018-2015
- **G = -2:** 2017-2015, 2018-2016
- **G = -1:** 2016-2015, 2017-2016, 2018-2017
- **G = 0:** 2015-2015, 2016-2016, 2017-2017, 2018-2018
- **G = 1:** 2015-2016, 2016-2017, 2017-2018
- **G = 2:** 2015-2017, 2016-2018
- **G = 3:** 2015-2018

When evaluating models, we aggregate performances for the different combinations of years pertaining to each gap by averaging them, i.e. performance scores for G = 0 would consist of the average of four combinations of years. We note that this setup inevitably leads to different numbers of experiments for each temporal gap with a smaller number of held-out testing sets for larger gaps.

In our evaluation, we use the following metrics: (1) **Model Evaluation Metric** which measures the model performance (macro-averaged F1-score) for each training and test pair; (2) **Performance Change Metric** which

**ALGORITHM 1**
Temporal Performance Evaluation

**Input** longitudinal training sets, development sets and test sets, $D_{tr}$, $D_{dv}$and $D_{ts}$ for each dataset $D$ covering $N$ temporal years, language representations $R$ and classification algorithms $C$.

**Output** Average performance score for all possible time gaps between training source years $i \in I$ and test target years $j \in J$ where ($I \subset N$ and $J \subset N$). ▷ we use N for both training source years $I$ and test target years $J$ as they are equal sets of all $N$ years in our settings.

**for** all $m \in Models^{R\&C}$ **do**
  **for** all training sets $D_{tr}$ from target year $i \in N$ **do**
    Train $m$ using $D^i_{tr}$ to get a classifier $m^i$
    Apply early stopping strategy using held-out $D^i_{dv}$
    **for** all testing sets $D_{ts}$ in $j \in N$ **do**
      Predict classes for $D^j_{ts}$ using $m^i$.
      Calculate the **Temporal Gap** (See Equation 1) to the temporal distance between training and test set (j - i) to determine the temporal distance of classifier $m^i$ predictions of $D^i_{tr}$ set.
      Calculate **Model Evaluation Metric** (See Equation 2) to Evaluate $D^j_{ts}$ performance of the given classifier using classifier $m^i$ predictions of $D^i_{tr}$ set.
    **end for**
  **end for**
  Compute **Performance Change Metric** for all years per time gap (See Equation 3).
**end for**

quantifies the difference between model performance (macro-averaged F1-score) on the testing set from the same year as training and on the testing set distant in time.

Our experimental setup is summarised in Algorithm 1, showing the procedure for aggregation of year pairs by their temporal gap.

- **Model Evaluation Metric:** For each training ($D^i_{tr}$) and test sets ($D^j_{ts}$), we use the macro F1-score (F-macro) measure to assess the performance between a single pair of training and test sets for any given model design choice. F1 score is a weighted average of the recall and precision.

$$F - macro(D^i_{tr}, D^j_{ts}) = \frac{2 \cdot precision \cdot recall}{precision + recall} \tag{2}$$

- **Performance Change Metric:** We quantified the performance fluctuation for each time gap by averaging F-Macro over all training and test pairs with the same temporal distance. We used an evolving longitudinal benchmark with multiple experiments for each time gap to accurately measure the temporal performance drop for each task.

$$P(G) = \overline{\sum_{i \in N} \sum_{j \in N} F - macro(D^i_{tr}, D^j_{ts})} \text{ if} j - i = G \tag{3}$$

## 4. Longitudinal datasets

In this section we introduce the datasets used for our experiments and perform a preliminary analysis of their characteristics.

### 4.1. Selected datasets

We chose three datasets for our study based on their longitudinal nature, i.e. they cover the span of several years, where each year is dense containing sufficient data for training and testing, and they are labelled for text classification tasks:

| Datasets | GESD | TESA | ABRR |
|---|---|---|---|
| **Classification Labels** | Binary: favour/against | Binary: positive/negative | Binary: 5/1 |
| **Labels balance** | unbalanced | balanced | balanced |
| **Annotation** | Aggregated hashtags | Aggregated emojis | Human annotation/tags |
| **Time range** | 2014-2019 | 2013-2020 | 2000-2018 |
| **Size per year** | 48,000 | 99,800 | 26,200 |
| **Total size** | 288,000 | 798,400 | 497,800 |
| **Sampling method** | GetOldTweets API | Twitter stream archive | - |
| **Source** | Twitter | Twitter | Amazon |
| **Data specificity** | Gender Equality | Generic | Book Reviews |

**Table 2**
Summary description of the selected datasets.

| | Source | | Target | Labels | |
|---|---|---|---|---|---|
| | **Train.** | **Dev.** | **Test.** | **% Class 1** | **% Class 2** |
| **GESD** | 35,100 | 3,900 | 9,000 | 76.9% (Support) | 23.1% (Oppose) |
| **TESA** | 74,850 | 9,980 | 14,970 | 50% (Pos) | 50% (Neg) |
| **ABRR** | 19,650 | 2,620 | 3,930 | 50% (5) | 50% (1) |

**Table 3**
Dataset statistics (per year) for train, development and test sets.

- **Gender Equality Stance Detection (GESD).** This is a dataset that spans from 2014 to 2019, with a collection of tweets related to gender equality, which were retrieved through the Twitter API based on a collection of relevant hashtags. The labelling of the dataset is done through distant supervision, by using a collection of opinionated hashtags which determine the stance of tweets (Alkhalifa et al., 2021). Tweets in this dataset are labelled indicating their stance as one of “support” or “oppose”.
- **Temporal Multilingual Sentiment Analysis (TESA).** This is a sentiment analysis dataset labelled through distant supervision for the 7-year period ranging from 2013 to 2020. The longitudinal sentiment analysis dataset is labelled based on a manually curated list of emojis and emoticons (Yin, Alkhalifa, and Zubiaga, 2021). Tweets are labelled for sentiment as either “positive” or “negative”.
- **Amazon Books Rating Reviews (ABRR).**[1] This dataset includes book reviews for the period from 2000 to 2018 (Ni, Li, and McAuley, 2019), which includes review ratings assigned by the users themselves along with the review texts. We used two columns from the dataset, **text** of the review and **overall rating** of the product. The original ratings in the dataset range from 1 to 5; in order to frame the task as a binary problem, we sample the reviews rated as either 1 or 5.

Two of the datasets come from social media (GESD and TESA), whereas the other one (ABRR) is made of user-generated contents in a more constrained setting (i.e. book reviews). Table 2 shows a summary of the datasets, including the time covered, size, annotation methodology as well as the task.

***Data sampling.*** For each dataset, we sample the same amount of data and preserve the same distribution of labels for each year, which helps us avoid other confounding factors to solely focus on the impact of temporal change. In all the settings, we applied a stratified split with 75%, 10%, 15% for train, development and test per year respectively. The resulting distribution of data is shown in Table 3.

## 4.2. Analysis of the temporal dynamics of language use

We look at the word frequency in the datasets over time, i.e. how does word usage persist over time and to what extent is word usage ephemeral. Figure 2 shows the statistics for different types of words according to their lifetime, which we define as follows:

[1] http://jmcauley.ucsd.edu/data/amazon/

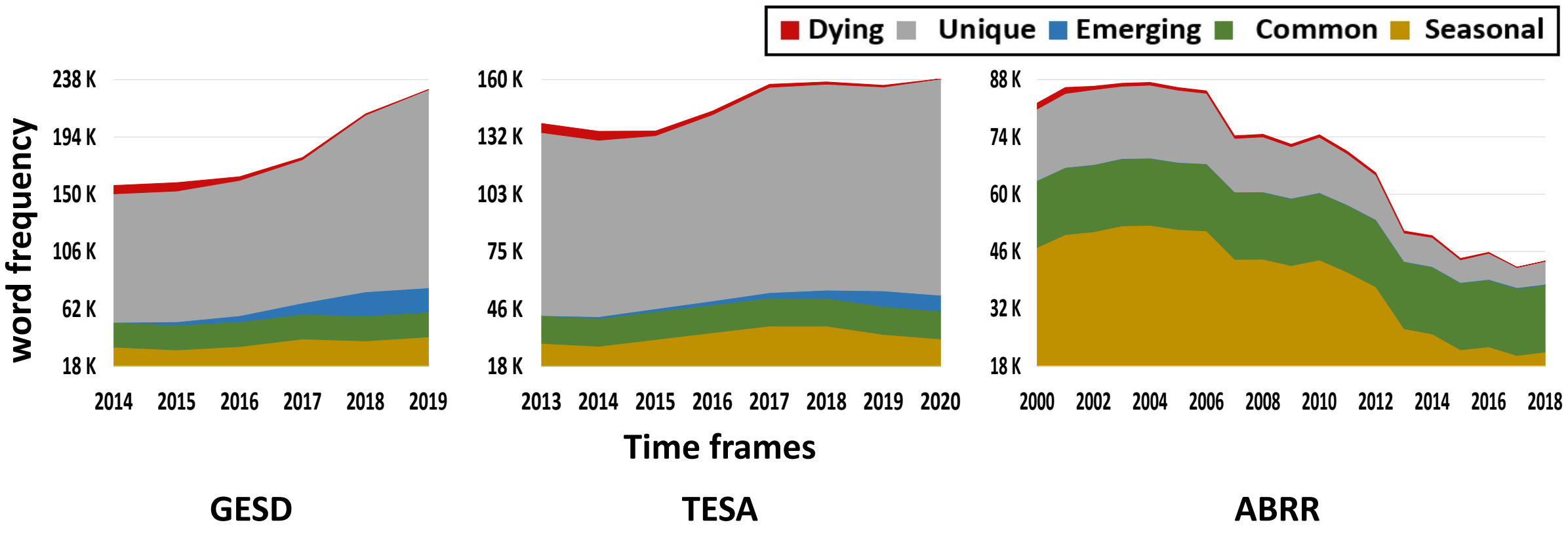

**Figure 2:** Temporal usage of different word types. Dying words (red), unique words (gray), emerging words (blue), common words (green) and seasonal words (brown). See main text for descriptions.

- **Dying words (red):** Words that, having been used for 2+ years, are no longer used in future years.
- **Unique words (gray):** Words only used in that year and not in any other year.
- **Emerging words (blue):** New words in that year which had not been used in the past.
- **Common words (green):** Words consistently used in all the years. Note that the set of common words is the same across all years, and hence absolute values of common words do not change.
- **Seasonal words (brown):** Words used in 2+ years but not in all the years.

The patterns observed in this analysis lead to the following hypotheses for our experiments:

- **Total number of words.** We observe an increase of vocabulary size for the social media datasets (GESD and TESA), and a decrease for the ABRR dataset of reviews. This may be due to the nature of the data, i.e ABRR coming from a more constrained domain (book reviews), whereas the social media domain attracts more diverse participants and is more dynamic and informal. The vocabulary increase is particularly expected for TESA, which is not restricted by a particular domain and covers a broad range of topics. We hypothesise that these patterns in word counts may have an impact on temporal model performance.
- **Emerging and dying words.** We observe that the proportions of emerging and dying words in the vocabulary of the ABRR dataset are very small. This suggests that the constrained nature of book reviews leads to a more established vocabulary, which varies to a lesser extent between years. However, emerging and dying words are more prominent in social media datasets (GESD and TESA). This shows a higher vocabulary variation in these datasets, which could lead to challenges for the models to classify instances from these datasets over time. While one may initially expect a higher variation in the TESA dataset due to its domain independence, the percentages of emerging and dying words are especially high for the GESD dataset. We believe that the larger variation in GESD can be explained by the evolving nature of the gender equality domain, where people's opinions are likely to evolve over time, leading to a larger vocabulary variation. Hence, we hypothesise that GESD might be a dataset (and a task setup) where models might more prominently drop in performance over time.
- **Unique and seasonal words.** We observe that the proportion of words of ephemeral nature, such as unique and seasonal words, does not vary much in the social media datasets TESA and GESD. Unique words are frequent, covering approximately 60% of words in the vocabulary, whereas seasonal words are rare. Interestingly, ABRR shows a very different trend. Seasonal words are more frequent than unique words, with both showing a decreasing tendency over time. The decreasing vocabulary size over time in ABRR is likely a contributing factor, i.e. a smaller vocabulary over time leads to more common words and fewer ephemeral words.

- **Common words.** The set of common words used consistently across all years is larger for ABRR than for GESD and TESA. This is remarkable given that ABRR is the dataset covering the longest temporal period (19 years), which reduces the likelihood of words to consistently occur annually over such a long period of time. However, we believe that the more restricted nature of book reviews leads to such vocabulary consistency.

The language dynamics observed in this analysis, along with the evident differences between datasets, highlight the challenging nature of building classification models that perform accurately over time, which in turn motivates the need for studying the temporal persistence of classifiers.

## 5. Language models, classification algorithms and methods for lexical analysis

We perform experiments with a wide range of state-of-the-art static and contextual language representations, and neural classification models to evaluate by-design adaptability to temporal changes. The feature space is always generated using word-level vectors extracted from the selected set of pretrained language models, including static and contextual representations, and the model parameters are learned from the training data, and hence can also be impacted by the quality and temporal persistence of the Pretrained Language Model (PLM) in question.

### 5.1. Pretrained language models

We experiment with two types of word representations, with three different methods of each type. The models are selected with the aim of having a diverse, competitive and widely used set of word representation methods.

***Static Word Representations (SWRs):***

- **Google Word2Vec (G-W2V)** (Mikolov, Sutskever, Chen, Corrado, and Dean, 2013): G-W2V widely used word embedding model, trained on roughly 100 billion words from Google News data.[2]

- **FastText (FT)** (Mikolov, Grave, Bojanowski, Puhrsch, and Joulin, 2018): FT has 300 dimensional vectors trained from English Wikipedia by using the skip-gram method Bojanowski, Grave, Joulin, and Mikolov (2017).[3]

- **Twitter Glove (Glove)** (Pennington, Socher, and Manning, 2014): Glove is trained on the non-zero entries of a global word-word co-occurrence matrix. The model we use is trained from 2 billion tweets.[4]

***Contextual Word Representations (CWRs):***

- **Bidirectional Encoder Representations from Transformers (BERT)**[5] (Devlin, Chang, Lee, and Toutanova, 2018). Google trained BERT in 2018 out of 11K unpublished books and English Wikipedia articles, using 110M parameters. A static masked language modelling (Static-MLM) loss objective was used to fine-tune BERT, with this masking is done only once as part of the preprocessing step prior training. BERT is trained as a non-auto-regressive model that generates contextual representations from all previous and next tokens in a text, then generates representation output all at once. WordPiece (Wu, Schuster, Chen, Le, Norouzi, Macherey, Krikun, Cao, Gao, Macherey et al., 2016) tokeniser, a greedy approach for dividing words into smaller tokens during training, was used to produce these tokens from text. When used in downstream task, each out-of-vocabulary (OOV) word is split into known sub-words. BERT's vocabulary contains 30,522 words. This reduces ambiguity in sentence meaning and chances of losing important signals due to OOVs.

- **Robustly Optimized BERT Pretraining (RoBERTa)**[6] (Liu, Ott, Goyal, Du, Joshi, Chen, Levy, Lewis, Zettlemoyer, and Stoyanov, 2019). RoBERTa inherits many of BERT's capabilities, with improved pretraining efficiency. It is trained from the same data as BERT, in addition to the CC-News dataset. RoBERTa uses Byte-Pair Encoding (BPE) (Shibata, Kida, Fukamachi, Takeda, Shinohara, Shinohara, and Arikawa, 1999) for data compression, in a similar fashion to BERT's WordPiece tokeniser. During the pre-tokenisation step, the

[2] https://code.google.com/archive/p/word2vec
[3] https://fasttext.cc/docs/en/pretrained-vectors.html
[4] https://nlp.stanford.edu/projects/glove/
[5] https://huggingface.co/bert-base-uncased
[6] https://huggingface.co/roberta-base

WordPiece algorithm selects symbol pairings that increase the probability of the training corpora, whereas the BPE algorithm selects the most common combinations. The number of words in the vocabulary is 50,265. The model is trained with a loss objective based on dynamic masked language modelling (Dynamic-MLM). This distinguishes it from BERT in that it changes the masked word for the same sentence at each training epoch. Similar to BERT, the model accepts tokenised inputs ranging from 3 to 512 and produces 12 neural layers with 768 hidden units.

- **Generative Pretrained Transformer 2 (GPT)**[7] (Radford, Wu, Child, Luan, Amodei, Sutskever et al., 2019). GPT-2 was introduced in February 2019 by OpenAI, as a model trained on a corpus of 8 million web documents. Like RoBERTa, it employs BPE with space tokenisation. It differs from BERT and RoBERTa in that it is an auto-regressive model that generates outputs iteratively and predicts tokens unidirectionally based on their context through reading tokens from left to right. Unlike prior models, GPT-2 is trained with a causal language modelling (CLM) loss objective since it is intended primarily to create human-like writing in text generation tasks. Furthermore, it employs decoder attention blocks from the transformer design, as opposed to BERT, which uses encoder blocks. The vocabulary size is 50,257. GPT-2 accepts tokenised inputs ranging from 3 to 1024 and produces 12 neural layers with 768 hidden units.

## 5.2. Machine learning models

We use a set of **classification models,** including two types of classifiers, and three algorithms of each type of classifier. All architectures were chosen based on their robustness in many state-of-art text classification tasks.

### *Traditional classifiers:*

- **Linear Support Vector Machine (LinearSVC)** (Joachims, 1998). LinearSVC is an algorithm that searches for a hyperplane (decision boundary) and linearly separates classes, and is used in numerous NLP classification problems. The introduction of kernel functions allows LinearSVC to generalise to high dimensional and sparse feature spaces. Therefore, feature engineering is not required even when the number of dimensions exceeds the number of data samples. Moreover, it is able to handle large numbers of features.

- **Logistic Regression (LogisticRegression)**. A supervised classification approach, it is used in binary classification because it fits a single line to split the space exactly into two using a sigmoid function. Using a probability curve to predict classes makes LogisticRegression different from LinearSVC, and more similar to deep learning models. The LogisticRegression classifier can interpret model coefficients (Bruin, 2011), an effective indicator of feature importance.

- **Multinomial Naive Bayes (MultinomialNB)**. MultinomialNB is a probabilistic learning algorithm that assumes features to be conditionally independent. Its probabilistic nature makes it vulnerable to data sample distributions but robust to deal with infrequent features, unlike LogisticRegression and LinearSVC.

### *Deep learning models:*

- **Convolutional Neural Network (CNN)** (Kim, 2014). Despite its capacity to retain the order of feature elements, CNN is a shift-invariant classifier. CNN may process the entire text or a portion of it by taking reduced signals from the sequence and sliding over it with a single conventional head. A deep CNN may include additional layers, allowing it to decrease signal overruns for each phrase. CNN can also learn several levels of structured n-gram co-occurrences. This is mostly owing to its mapping during the reduction phase via a filtering mechanism with a certain size and sliding window. This enables the CNN to compress the feature space of words into smaller latent feature representations.

- **Long-short Term Memory network (LSTM)** (Hochreiter and Schmidhuber, 1997). With LSTM the information persistence is high (Salton and Kelleher, 2018) due to the use of recurrent units connected as chains and temporal backpropagation. This allows a classifier to learn based on long term dependencies between sequences regardless of textual length or dataset size. Moreover, the ability of LSTM to model temporal historical dependencies makes it useful to capture subtle semantic changes (Elman, 1990). The is done though temporal memory sequencing for modeling each sentence word by word while training the model. This allows LSTM to reveal the latent syntactic and semantic long term signals for any word even from noisy data. Thus, it generalises better than other non-memory dependent models. This is different than classical models which are usually trained to reduce the mean square error by considering current inputs only and without using memory units to keep historical states of inputs.

[7] `https://huggingface.co/gpt2`

- **Hierarchical Attention Network (HAN)** (Augenstein, Rocktäschel, Vlachos, and Bontcheva, 2016; Li, Xu, and Shi, 2019). The HAN architecture we use includes GRU units and an attention mechanism. The first layer is a bidirectional GRU (Cho, van Merrienboer, Gülçehre, Bougares, Schwenk, and Bengio, 2014) layer, which is a kind of RNN and an LSTM variant. GRU, on the other hand, is less complex and hence quicker than LSTM. The next layer is an attention mechanism between word vectors and sentence representations. Text classifiers can generalise prediction by identifying the latent properties that exist between individual word vectors and the average representations of all words in a sentence.

## 5.3. Preprocessing and classifier hyperparameters

We preprocess the inputs to maximise the coverage of words in the training set, following two steps to clean the dataset and to tokenise the words to match the SWR. Out-of-Vocabulary words in the training set added with zero vectors to SWR. The sentence inputs are padded and truncated to 128.

We use a softmax layer to obtain class probabilities as the final layer in all our classifiers' architectures. We also use early stopping strategy with restoring best weights by measuring the validation loss on same-period training data. The number of training epochs is set to 25 with the same hyperparameters.[8] Reported experiment results are the averages of three runs each.

Each pretrained language model's text classifier is a feed-forward neural network architecture in which word-level representations are used as initial weighted layers for input tokenised sentences. Then fed into a flattened layer followed by a final sigmoid layer that outputs class probabilities.In the case of RQ2, addressed in Section 6.2, all traditional classifiers used the default parameters and TF-IDF features. All deep leaning models use the Adam optimiser with the learning rate fixed at $2e^{-5}$.

## 5.4. Methods for lexical analysis

Along with our experiments, we perform lexical analyses of the datasets to support our findings, particularly for RQ4 and RQ5. To do this, we propose several measures looking into aspects that can explain the causes of performance patterns we observe in our experiments. We categorise these measures in two groups: (1) lexical variations across datasets where we propose word-level statistical measures across training and test sets, and (2) contextual variations over time where we propose a model to track context-level meaning drift over time using cosine similarities and variance.

### *5.4.1. Lexical variations across datasets*

To address **RQ4** in Section 6.4, we calculate various unsupervised metrics estimating the similarity between training and testing sets. We then analyse whether these metrics are able to predict how well a model will perform on the new test set. We calculate the Pearson correlation coefficient between the following four statistical measures and model performance to quantify their potential impact on model performance. Table 4 shows a summary of all quantitative measures, which we discuss next. We discuss the results of this analysis in section 6.4.

- **Familiarity score:** This is measured by relying on two metrics extracted from two sets of words: (i) overlap ratio, as the proportion of words, which occur in both sets of words, and (ii) uniqueness ratio, as the ratio of words that occur only in one of the sets of words. The familiarity score is calculated by dividing the overlap ratio by the uniqueness ratio.

- **Jaccard Index (JI):** The JI is based on the intersection and union of words in two vocabularies. The JI is calculated as the size of the intersection divided by the size of the union. The resulting coefficient is asymmetric.

[8] As noticed that models performed similarly to one another, we assume that our findings apply to other hyperparameters; more study is required with consideration to computational costs.

| Measure | Definition |
|---|---|
| Familiarity score | $familiarity(U,V) = \frac{\|U \cap V\|}{\|V - U\|}$ |
| Jaccard index | $jaccard(U,V) = \frac{\|U \cap V\|}{\|U \cup V\|}$ |
| TFIDF similarity | $similarity(U,V) = tfidf(t,d,U) * tfidf(t,d,V)$ |
| Information rate | $H(V \mid U) = -\sum_{v,u} p_{V,U}(v,u) \log_b \frac{p_{V,U}(v,u)}{p_U(u)} = H(V,U) - H(U)$ |

**Table 4**
Summary of different evaluation measures used to quantify Lexical variations cross-datasets using unlabelled data. Note: $U$: vocabulary of training and $V$: vocabulary of test set.

- **TF-IDF:** By calculating the TF-IDF weights of words in two different vocabularies, we can then calculate the similarity between the two vocabularies. The similarity is calculated using the cross product similarity. This is also an asymmetric measure and avoids computation of sentence-by-sentence similarities.
- **Information rate:** By training a Markov model on a source vocabulary, and testing it on a target vocabulary, we rely on conditional entropy to estimate the information rate of the target vocabulary. The benefit of using information rate is to test the model's ability to predict the next word given the conditional probabilities from the training set. The information rate quantifies the amount of information required to describe the testing set $V$ given training set statistics $U$. With $p_U$ the probability distribution of $U$, and $p_{V,U}$ the distribution for the joint distribution $(V, U)$, the base- $b$ conditional entropy.

### 5.4.2. *Temporal language variations*

To answer **RQ5** we aim to analyse language variations over time by looking at the context that is surrounding the aspects extracted from the corpora. In this case, we choose to work with aspects that capture opinionated multi-word expressions, owing to two reasons: (i) given the nature of our datasets (sentiment, stance, reviews), to analyse how context surrounding these core elements changes, and (ii) because aspects are less likely to be polysemous than isolated tokens, hence reducing the potential of noise from multiple co-existing meanings in the analysis. Year-specific discrete representations of aspects can help us analyse how their context changes over time. To achieve this, we develop a novel approach to generate time-specific representations of aspects, allowing to quantify their change over time based on how the surrounding context changes.

Our approach to measure the temporal change of context is illustrated in Figure 3. It consists of two main steps: (i) Aspect Extraction Pipeline, where we identify aspects present in texts, and (ii) Contextual Change Measurement Model, where we measure how much the context of these aspects has changed.

**Step 1: Aspect Extraction Pipeline:** We first extract aspects, or multi-word expressions, from texts. To extract aspects, we rely on two resources:

- A lexicon of 7K words we build from two source dictionaries: (1) opinion lexicon (Biber, 2006) including adverbs and adjectives of different opinion modalities which covers different semantic categories: possibility, necessity, certainty, likelihood, attitude, communication, evaluation (210 lexicon words); (2) polarised sentiment lexicons (Liu, Hu, and Cheng, 2005) (2K positive and 5K negative lexicon words), which also covers misspelled words.
- The Spacy part-of-speech (POS) tagger (Hu and Liu, 2004; Honnibal and Montani, 2017), which we use to tag the texts.

Next, we look for sequences of adjacent words that both (i) contain a word present in the lexicon, and (ii) based on the output of the POS tagger, match one of the 45 POS regular expressions defined by (Hu and Liu, 2004). The set of sequences matching both criteria constitute our final set of aspects.

By analysing the frequencies of the resulting aspects over time, we can categorise them based on their use frequency over time into one of dying, unique, emerging, seasonal or common, following the same approach as in Section 4.2. Figure 4 shows the result of the analysis, demonstrating that dynamic aspects are frequent and their use varies over time.

**Step 2: Contextual Change Measurement Model:** The change measure model consists in turn of the following three steps: (1) generate time-specific embeddings of aspects, (2) measure pairwise similarities between time-specific embeddings of an aspect over time and (3) rank aspects by their contextual change.

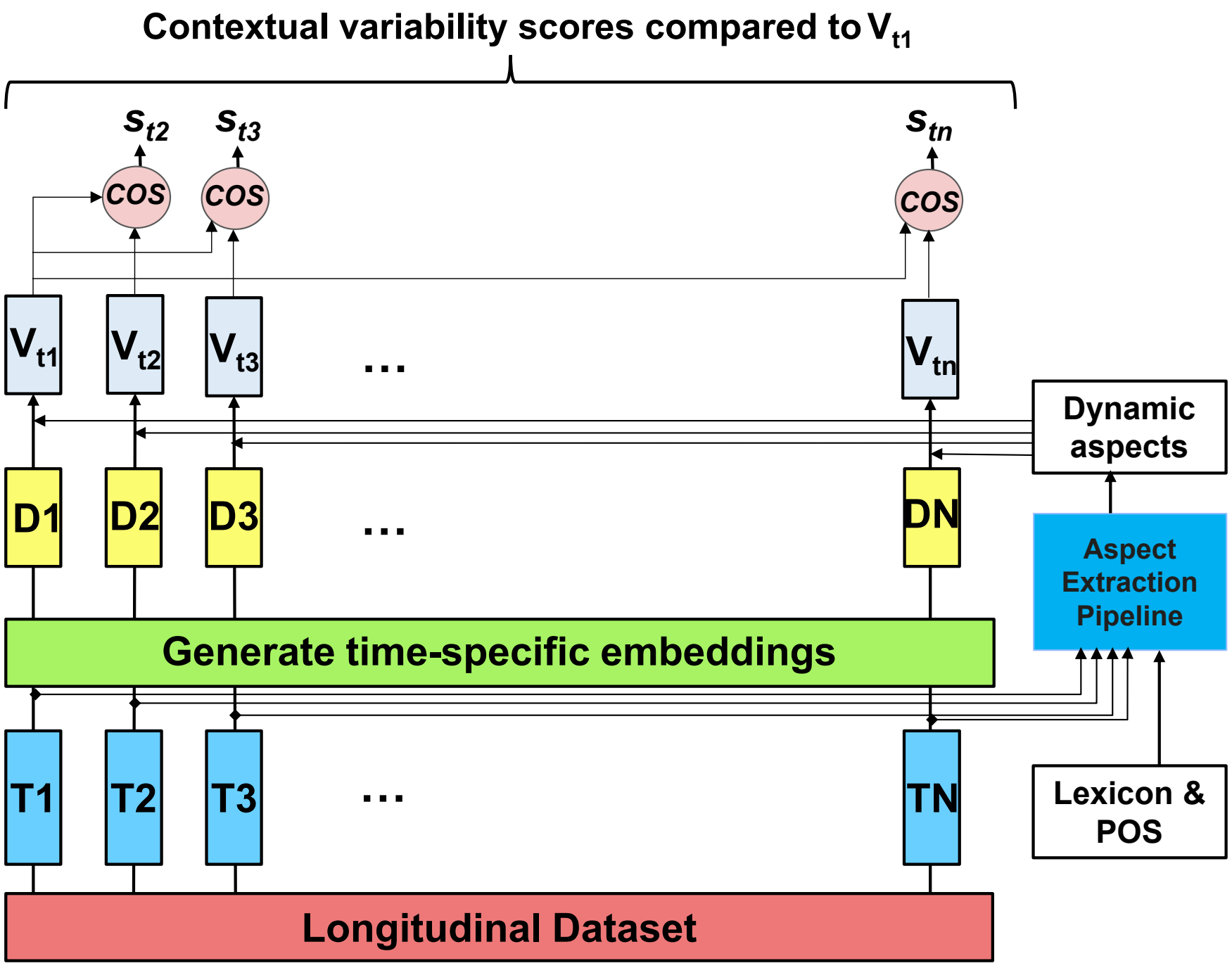

**Figure 3:** Our approach for assessing contextual coverage for dynamic aspects over time. The dataset is first split into years. For each year, we then pretrain a contextualised model (e.g BERT, RoBERTa) using a masked language modeling strategy. Having this year-specific language models, we can then calculate similarity scores across years. When these similarity scores are low for a particular aspect, they indicate a prominent change in its surrounding context over time.

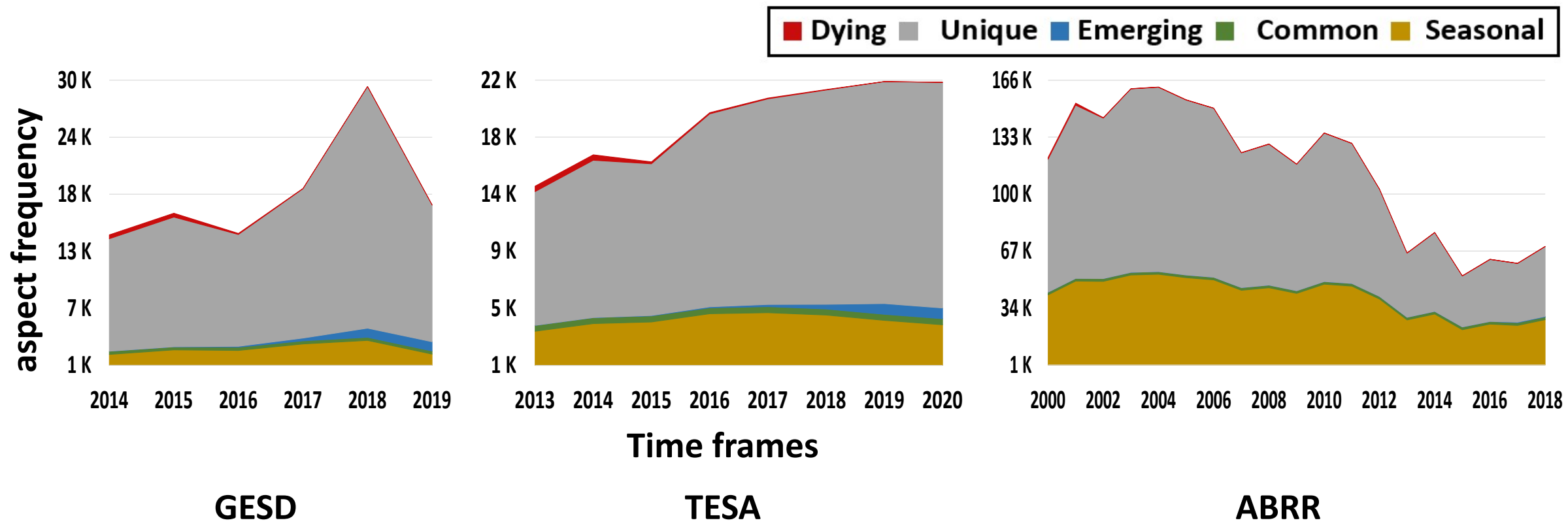

**Figure 4:** Temporal usage of different aspect types. Dying aspects (red), unique aspects (gray), emerging aspects (blue), common aspects (green) and seasonal aspects (brown).

1. **Step 2.1: Generating time-specific embeddings:** Given a longitudinal dataset $D$ that spans several years $\{T_1, T_2, \dots, T_N\}$ where each year has the same number of sentences per year (SpY), and a context-based PLM. A discrete set of time-specific embedding can be trained using masked language strategy (MLM) for each year. We obtain 6 discrete time-specific embeddings for GESD dataset, 8 for TESA and 19 for ABRR.
2. **Step 2.2: Measuring similarities between two time-specific representations of an aspect:** The contextual change of an aspect from time $T_1$ to each time $T_i$ can be calculated by measuring the cosine similarity between the relevant time-specific representations (Hamilton, Leskovec, and Jurafsky, 2016; Shoemark, Liza, Nguyen,

Hale, and McGillivray, 2019; Tsakalidis, Bazzi, Cucuringu, Basile, and McGillivray, 2019). The inverse of the cosine simlarity then indicates contextual change, i.e. lower cosine similarity indicating bigger change. In the interest of simplicity and focus, similarity scores are computed for each year with respect to year $T_0$, which is kept constant as a pivot.

$$s_{ti}(a, D1, Di) = \cos(V_{T1}, V_{Ti}) = \frac{V_{T1} \cdot V_{Ti}}{\|V_{T1}\| \|V_{Ti}\|} = \frac{\sum_{j=1} v_j^{T1} v_j^{Ti}}{\sqrt{\sum_{j=1} {v_j^{T1}}^2} \sqrt{\sum_{j=1} {v_j^{T1}}^2}}, \tag{4}$$

where $V_{T1}$ and $V_{Ti}$ are the embeddings of aspect $a$ for years $T_0$ and $T_i$, trained using datasets $D1$ and $Di$, respectively. We repeat this process several times for all $i \in \{t+1, ..., T\}$ for all $N$ temporal data frames in a given longitudinal dataset $D$. For each aspect, we then get a final list of cosine similarities $s$ between $T_1$ and all $T_i$.

3. **Step 2.3: Ranking aspects by contextual change:** To measure the contextual variability of target aspects over time, we use the variance ($\sigma^2$) to measure the fluctuation of cosine similarity scores from the mean similarity score ($\mu$) over time of the same aspect, defined as follows:

$$\sigma^2 = \frac{\sum_{i=1}^{N} (\cos(V_{T1}, V_{Ti}) - \mu)^2}{N} \tag{5}$$

## 6. Experiments, results and analysis

In this section, we present the results of the five experiments we conduct to address the five research questions we set forth. We first assess the extent to which different language models (RQ1, Section 6.1) and algorithmic architectures (RQ2, Section 6.2) persist their classification performance. Then, by looking at temporal gaps (RQ3, Section 6.3), linguistic features (RQ4, Section 6.4), and the similarity of a given set of dynamic aspects contextual variations (RQ5, Section 6.5), we concentrate our studies on the linguistic complexity through temporal prospective to determine when and how diachronic changes impact text classifiers' generalisability.

### 6.1. Experiment 1: Impact of language representations on performance (RQ1)

In this experiment, we aim to answer RQ1 to assess the impact of language representations on the classification performance. We use the six different language models presented in Section 5.1. We also present grouped performance averages for each group consisting of static (SWRs) or contextual (CWRs) word representations.

Classification results for different language representations across the three datasets are shown in Figure 5. Most importantly, we observe performance decay for all datasets and language models at least in one direction (future or past) comparing to the initial temporal gap of 0 (present). Moreover, the performance decay for each PLM and dataset remains consistent, where all PLMs show similar decaying trends. Despite these commonalities, we make some interesting observations:

- The performance decay is not the same for all datasets. We observe that the decay is particularly steep for the GESD dataset, where even the top-performing representation (BERT) drops performance by more than 15% for data 5 years in the past and by more than 10% for data 5 years in the future. This drop is comparatively modest for the other two datasets, TESA and ABRR. In the case of TESA, we observe that all representations drop slightly in performance for past and future data, with the exception of BERT that achieves some degree of stability for the past data. The trend line is different for the ABRR dataset, which shows performance stability on future data, likely due to the shrinking nature of its vocabulary over time, as shown in Section 4.2.

- SWRs exhibit a substantial drop in performance for the past data, and lesser drop for the future data; the more pronounced drop of SWRs compared to CWRs is likely due to the ability of the latter to model context, which enables some readjustment of words that change meaning over time thanks to information extracted from the surrounding context. To better understand and explain these different patterns we observe across datasets, we will delve into dataset characteristics in RQ3 (section 6.3).

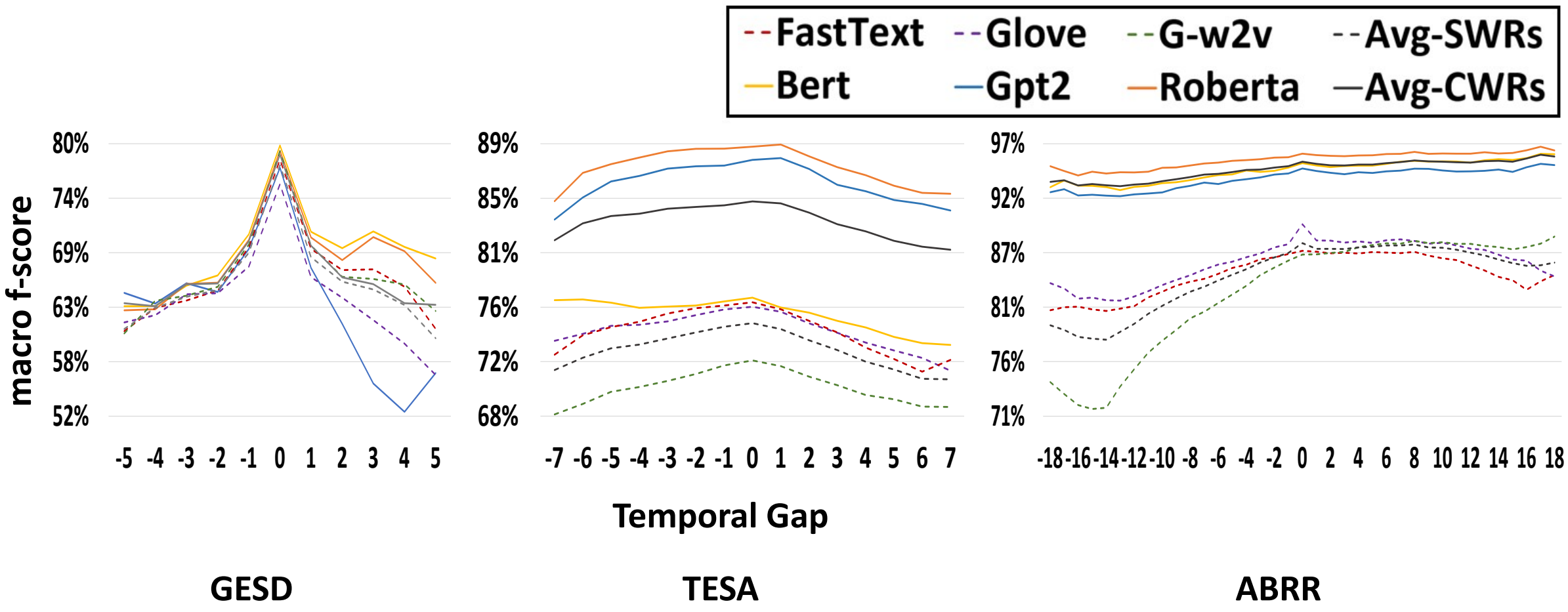

**Figure 5:** Temporal performance of different language representations across the three longitudinal datasets. *Dashed-line*: average f-score results for average static-based representations, *Bold-line*: average f-score results for all representations for average context-based representations.

- Looking at the absolute performance scores of different representations, we observe that CWRs consistently outperform SWRs, which is in line with the expectations. However, we do observe some variation across the different CWRs, as for example BERT performs best for GESD, but its performance leaves much to be desired for TESA. We observe an overall best generalisation across datasets for the RoBERTa model, which performs best for TESA and ABRR, and close to the best for GESD. The improved generalisability of RoBERTa over BERT may be due to its larger pretrained vocabulary, as RoBERTa is trained from the same corpora as BERT, with the addition of CC-News.

## 6.2. Experiment 2: Algorithmic architecture impact on performance (RQ2)

In this experiment, we address RQ2, assessing the extent to which different algorithmic architectures enable temporal persistence (please refer to Section 5.2 for the full description of the algorithms used). In the interest of focusing on algorithm impact, we average performances for each type of language representation, i.e. average of SWRs and average of CWRs.

Figure 6 shows the temporal performance scores for the different classification algorithms under study. We observe the following:

- The overall tendency is for all models to exhibit similar performance trends over time. Despite some models achieving better absolute performance than others, the performance drop across models is largely consistent. The performance drop we observe for future and past periods is mostly symmetric for GESD and TESA, where the performance drop increases for larger temporal gaps. ABRR shows a similar trend for past periods, however with a different pattern for future periods, where performance remains stable or even increases slightly on occasions.

- In terms of absolute performance scores, we observe that HAN-based models achieve an overall better performance than other models, which however also suffers from the same performance drop pattern as the other models. This is consistent with the findings in RQ1, and hence shows limited impact of the classification algorithm when we look at temporal persistence as the key factor.

Similar to what we observed in the first experiment dealing with language representations, and leaving absolute performance scores aside, again we see predominantly consistent performance trends regardless of the algorithm choice. The shape of these performance trends is in fact very similar for different language representations and algorithms selected, which highlights the impact that the data itself has on this performance drop. While we have seen that some pretrained language models can achieve improved generalisability in temporal persistence (as is the case of RoBERTa thanks to its broader vocabulary than BERT), we cannot conclude the same for classification algorithms, as we do not see a similarly clear difference in this case. We can therefore that, when it comes to temporal persistence, language representations obtained through pretrained language models have a bigger impact than classification algorithms, which are not as crucial.

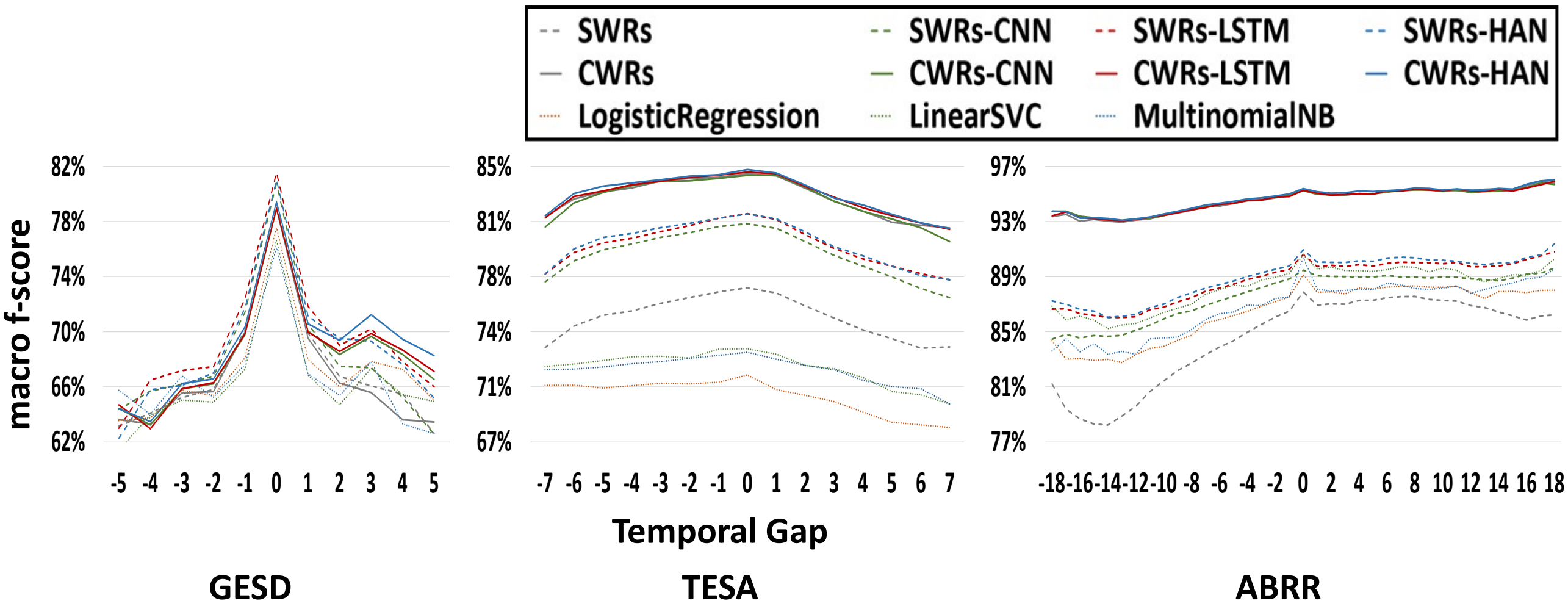

**Figure 6:** Temporal performance of different algorithmic architectures across the three datasets. *Dotted-line*: f-score results for each traditional machine learning model with frequency-based representations, *Dashed-line*: average f-score results for average static-based representations, *Bold-line*: average f-score results for all representations for average context-based representations.

Aside from pretrained language models and classification algorithms, looking at the different datasets we observe that the performance drop is more prominent and consistent for the TESA and GESD social media datasets. The trends are different for ABRR, which, as a dataset collected from reviews associated with a particular domain, i.e. books, one could hypothesise that leads to a lesser impact over time due to the predictably more constrained vocabulary. The clear differences in performance across datasets leads to our third research question, where we conduct more quantitative experiments to investigate our selected datasets structure to gain deeper understanding of models temporal performance fluctuation.

## 6.3. Experiment 3: How the temporal gap impacts performance (RQ3)

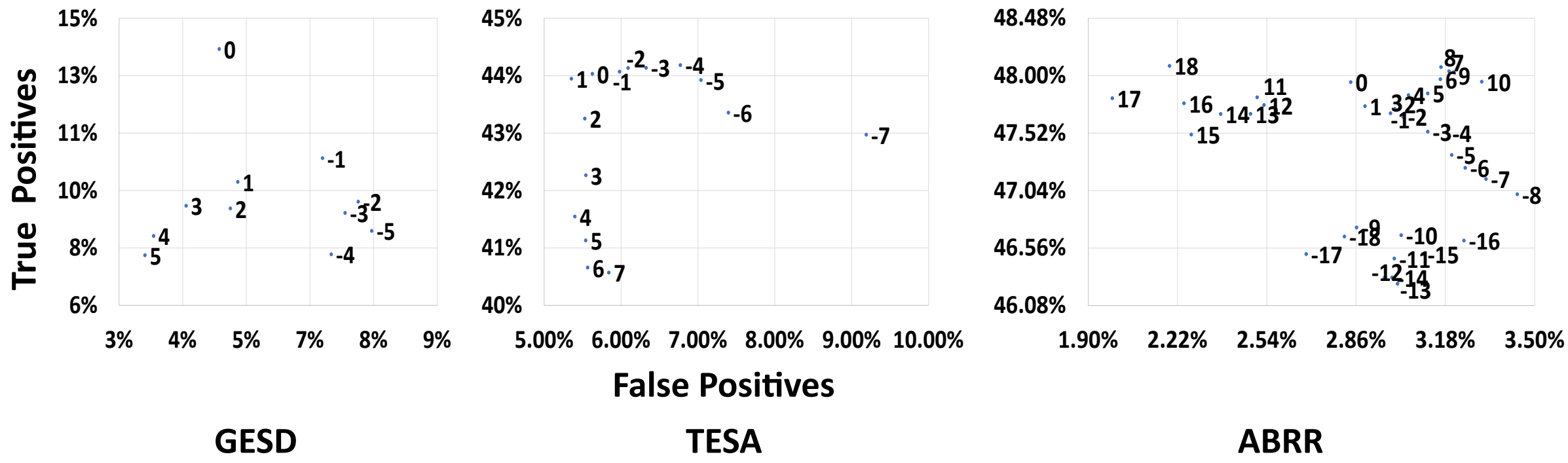

**Figure 7:** Model performance across temporal gaps. The best performing sets are the one positioned in top left corner of the grids, i.e. those minimising false positive rates and maximising true positive rates.

In this section we address RQ3 by looking at how different factors in the inherent characteristics of the datasets impact model performance over time. Figure 7 shows TP (true positive) and FP (false positive) rates as a scatter plot for different temporal gaps, where data points are labelled with the corresponding temporal gap.[9]. This allows us to rank temporal gaps based on TP and FP rates. Note that the optimal results are those minimising FP and maximising TP, hence values positioned in the top-left corner of the figures are ideal. In the interest of clarity, we focus here on results produced by CWRs-HAN model as the best-performing model in the previous experiments.

| Vocabulary Test/Dataset | *GESD* | *TESA* | *ABRR* |
|---|---|---|---|
| *Familiarity score* | +54% | +48.5% | +57.7% |
| *Jaccard index* | +73.3% | +29% | +65.5% |
| *TF-IDF similarity* | +49.1% | +46.8% | +14.1% |
| *Information rate* | +64.2% | +34% | +52.5% |

**Table 5**
Lexical variation correlation scores (and two-tailed p-values $< 0.05$ significant results) of metrics against model's performance at best overall prediction model (CWRs-HAN) for each dataset.

For both GESD and TESA, the best performance is achieved for the temporal gap 0, i.e. same-year train and test. In the case of GESD, we observe that performance is better for future tests, whereas performance is similar for future and past test sets for TESA. The trend is quite different for ABRR, where performance for future test sets is even better than for temporal gap 0, which confirms our finding in RQ2 showing that the constrained vocabulary in this dataset leads to improved performance. This makes future test sets less linguistically diverse, with better coverage of common words over time, facilitating classification of future data in ABRR. This can be observed in the cluster of larger temporal gaps (11+ years) in the top-left corner of the ABRR figure.

## 6.4. Experiment 4: How lexical variations across datasets help determine temporal persistence (RQ4)

With RQ4 we aim to investigate if there is a metric that we can use to estimate the temporal persistence (or lack thereof) that a dataset will exhibit, given the realistic scenario where we lack longitudinally labelled data. Can we estimate this temporal persistence for the linguistic characteristics of an unlabelled dataset and, if so, with what metrics?

To address RQ4, we analyse the Pearson correlation scores between four different factors in dataset vocabularies and model performances with CWRs-HAN model predictions (see Section 5.4.1 for implementation details).

Table 5 shows the resulting correlation scores for the four factors under study. A first look at the correlation scores shows consistent positive correlation scores across the board. However, the values of these positive correlation scores vary substantially, showing different degrees of strength in these correlations. The familiarity score is similarly high for all three datasets. In the case of GESD and ABRR, both the Jaccard index and the information rate are particularly high, whereas the TF-IDF similarity score is higher for both GESD and TESA than for ABRR. All in all, however, we could determine that, among the metrics under study, the familiarity score, given its consistency across the dataset, is the best metric to predict model performance deterioration. Indeed, where one lacks labelled data for years beyond the current year 0, one could estimate model performance drop by using the familiarity score as a metric.

To further understand the insights derived from the familiarity score, we delve into its results with finer granularity. Figure 8 illustrates pairwise familiarity score values for each pair of years. The first clear pattern difference we observe is between ABRR and the other two datasets. With ABRR, we see that the highest familiarity scores are between initial years for training (e.g. 2000) and final years for testing (e.g. 2018), a similarity that decreases even for same-year comparisons for early years (e.g. 2000) and for past year comparisons (e.g. 2018 training and 2000 testing). This observations through the familiarity score is in line with the performance scores observed in earlier sections for ABRR. On the contrary, both TESA and GESD show show higher familiarity scores for years that are closer to each other, which decrease as years are further apart from each other. This is again in line with the performance scores we have observed.

This again reaffirms the validity of the familiarity score as a metric to estimate the model performance deterioration for a particular dataset where we lack labelled data for more than a single year, i.e. we can calculate the familiarity score based on the overlap and uniqueness of the vocabulary sets in different years of unlabelled data, and we can expect some degree of correlation with the actual performance.

[9]Similar to ROC Curve (Flach, 2012), as we have data-centric approach for using same model we refrain from using same name

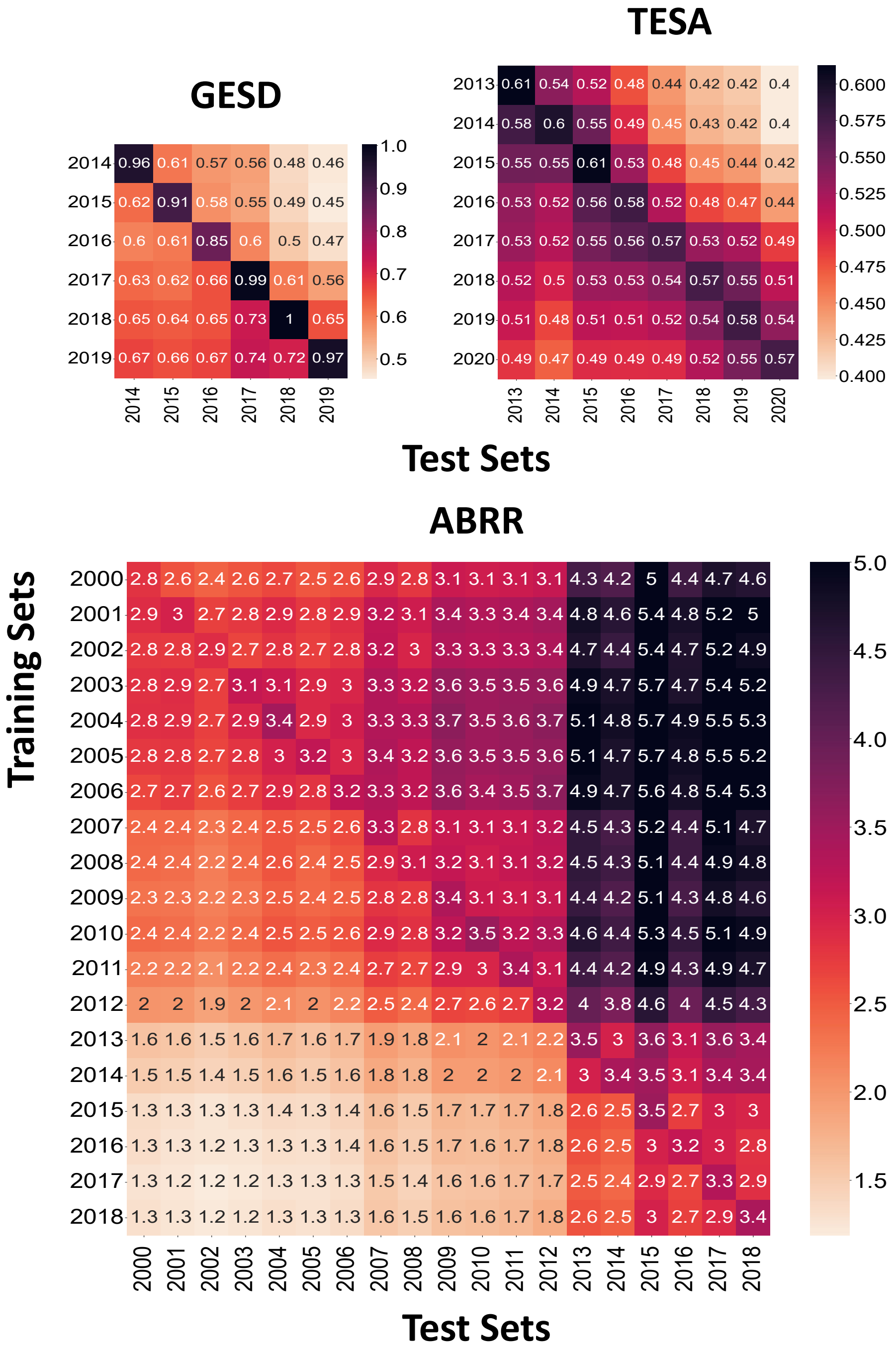

**Figure 8:** Temporal effect of familiarity score in each train-test pair. Darker cells indicate higher rates.

## 6.5. Experiment 5: Generalisability of contextual characteristics of context-based models (RQ5)

To address RQ5, following the methodology described in Section 5.4.2, we quantify how the changing context of unique aspects impact the resulting representations of contextualised language models. The aspects we use to measure these changes are short, multi-word expressions, such as:

***"@realDonaldTrump great"*** from TESA

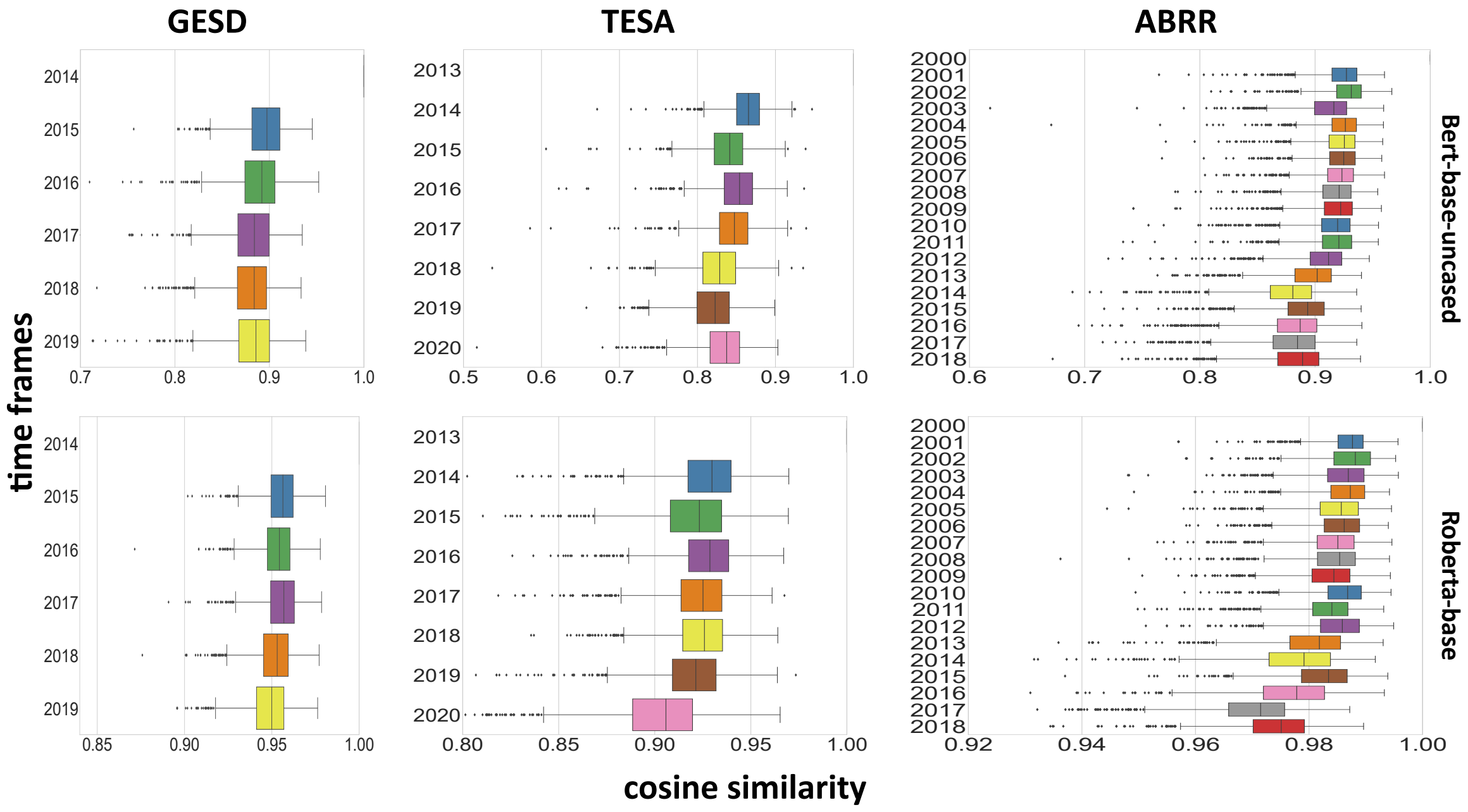

**Figure 9:** Assessing context-based temporal semantic similarity decay using discrete temporal context-based representations for BERT and Roberta by measuring the cosine similarity of dynamic unique aspects.

This is an example of an aspect occurring in only one of the years in the TESA dataset. For this specific example, we observe that its contextual similarity has dropped gradually as we move further away from the first year, with the following similarity scores: 2014 (0.8542), 2015 (0.7838), 2016 (0.6222), 2017 (0.6222), 2018 (0.6119), 2019 (0.5368), 2020 (0.5171). Using multiple examples like this, we can determine whether transformer-based models produce stable representations over time despite the contextual changes. To do so, we next assess whether the trend observed in the example above generalises to other aspects in the dataset.

Figure 9 shows the similarities of all aspects across years and across datasets. Each box plot represents the similarity scores for the set of aspects under consideration. The box plot for each year represents the similarities in that year with respect to the first year in the dataset, which is taken as a reference. For figures looking at common aspects, unique aspects and seasonal aspects, we observe a consistent tendency for the similarity scores to decrease over time, as the year in question is further apart from the first year.

This indicates that meanings and contextual representations keep varying over time, where this change does not happen all of a sudden, but it gradually happens showing a decreasing tendency of similarity scores. This in turn indicates that context-based language models do not fully capture the evolution of these aspects. Even if context-based language models provide the means to process sub-word, contextualised representations of texts, this has still room for improvement in furthering temporally stable representations to enable temporal persistence of models.

Indeed, learning a meaningful contextual representation for a unique aspect A at time T is much harder for models distant in time compared to models closer in time although A has never been seen in any other time-frame. While both BERT and RoBERTa showed slight temporal changes in the context when using MLM in which temporal splits are used to generate the discrete temporal embeddings. This change is smaller in RoBERTa compared to BERT which is potentially a result of the dynamic masking strategy for sentences utilised in RoBERTa. This again indicates, as discussed earlier in RQ1, that RoBERTa proves to be more stable than BERT when it comes to temporal persistence.

For more details showing specific examples of aspects and their scores, see Table A1 in Appendix A.

## 7. Discussion

Our work provides a first-of-its-kind study looking into both quantifying the challenges of developing temporally persistent text classifiers using dynamically evolving datasets, as well as further understanding the impact of dataset factors into this performance drop. Beyond simply confirming or quantifying that model performances drop over time, we further dig into getting a broader understanding into the problem by getting insights applicable in future experiments. We envisage the situation where one may need to choose a persistent classification strategy, but where longitudinally annotated datasets are lacking at the time of making this decision. Hence our overarching research question revolves around: what are the properties of the dataset, model or task at hand that can determine its temporal persistence?

A first look at the results from our experiments confirms that state-of-the-art language models and algorithms experience a performance drop over time as data evolves. This performance drop shows a similar trend compared to deep learning models, despite achieving overall better performance scores in terms of absolute values. This is especially true with user-generated content and social media datasets, where less formal and less established vocabularies are prone to evolve quickly. Both our experiments and our analyses demonstrate this in terms of performance scores and lexical changes, respectively. Where previous research has also posited the importance of capturing social changes, e.g. Syria and Iraq associated with war context (Kutuzov, Øvrelid, Szymanski, and Velldal, 2018), where new words emerge and word meanings evolve, there is a growing need to develop temporally persistent models.

Along with these general findings, a deeper look into the results, by investigating the five research questions we defined, leads to interesting and important findings that helps inform the design of text classifiers with their temporal persistence as the objective.

## 7.1. Summary of key findings and best practices for classifier design

We next summarise our key findings from this research, along with our suggestions for best practices for classifier design, which we highlight in bold.

Our initial vocabulary analysis of three longitudinal datasets shows varying patterns of different types of words. We observe that social media datasets (GESD and TESA) experienced an increase of vocabulary size over time, whereas the book review dataset (ABRR) experiences the opposite effect by exhibiting a decrease of vocabulary size. A possible explanation of this is that a more constrained domain such as book reviews may tend to settle into a fixed vocabulary, as opposed to more open domains. Having a more fixed, less varying vocabulary, also means having fewer emerging words, as well as fewer ephemeral words, both of which have shown to have a strong impact on model performance drop over time. Indeed, the reduced vocabulary size and variation leads to improved persistence of model performance over time in the case of ABRR.

Language models (RQ1) and algorithms (RQ2) both exhibit similar trends in their performance drops. This performance drop is particularly prominent for GESD, a dataset that deals with stance classification on the timely topic of gender equality, which is expected to have fluctuated over recent years. This means that the GESD dataset may be also impacted by how people expressed their opinions on this evolving topic, which in turn translated into a more prominent vocabulary change and a more noticeable performance drop in a shorter period of time (5 years) than in the other two datasets (7 and 18 years). Among the other two datasets, TESA experiences a larger performance drop than ABRR, again owing to the vocabulary stability and lack of vocabulary growth observed in the latter, as explained above.

Where all language models and algorithms show a similar trend in their performance decay, there is a relatively high consistency with the best-performing combination of model and algorithms in temporal gap 0 also performing best over time. Based on this observation, where one only has accessed to labelled data from year 0, **a wise approach to design a robust classifier can be based on the combination of language model and algorithm leading to top performance in year 0**.

Despite a consistent trend in performance drops across language models and classification algorithms, however, we observe that language models do exhibit different abilities to generalise better in terms of temporal persistence across datasets. This is particularly true when we compare RoBERTa and BERT, where we see that the former achieves better generalisability across datasets. Hence, despite having room for improvement in terms of temporal persistence, **when designing a persistent text classifier, it is safer to choose RoBERTa over BERT provided its improved generalisability across datasets**. One possible explanation is that the extended vocabulary of RoBERTa enables this improved generalisability, and therefore it opens an avenue for future research in further increasing its vocabulary. While we see this interesting difference between the two pretrained language models RoBERTa and BERT, we do not see a similar, informative difference between different classification algorithms, all showing similar trends in terms of persistence and generalisability.

When we look at time (RQ3), we observe that –predictably– the best model performance is generally achieved when the training and test datasets pertain to the same year (gap 0). However, surprisingly this is only true for two of the datasets (GESD and TESA), and we observe that performance scores can be higher when a model trained on older data is tested on future ABRR data, even when they are 18 years apart. Despite this unexpected outcome, it again highlight the nature of the ABRR dataset which has a very different vocabulary pattern than the other two datasets. A closer look at the characteristics of ABRR indeed shows a temporally shrinking vocabulary growth, indicating that the vocabulary in a more constrained domain like Amazon book reviews (compared to more informal, evolving social media data) makes it easier to achieve close to persistent performance. Hence, **an analysis of the vocabulary growth exhibited by a dataset is an important factor to consider when designing a persistent text classifier, with datasets showing low growth requiring less effort to achieve persistence**.

Further looking at correlations between linguistic features and model performance drop (RQ4), we find that the four linguistic factors we studied (familiarity score, jaccard index, similarity and information rate) play an important role in determining model performance over time. Indeed, one can effectively use these factors to estimate the temporal persistence of models. Where **one has labelled data from year 0 and unlabelled data from other years, a classifier design informed by these metrics can lead to a better decision**. More specifically, we have found that, among the metrics under study, the familiarity score provides the best estimate of model deterioration without using labelled data beyond year 0. However, where one only has labelled data from year 0, but **no further unlabelled data yet, one can instead attempt to estimate these metrics by relying on other more qualitative factors**, such as whether the data comes from social media and whether the domain contained in the dataset is expected to vary substantially.

Looking at the temporal persistence of contextual language models (RQ5), we found that while the emergence of new words can be solved by sub-word tokenisation in contextual-based embedding models, words dynamic between training and test pairs still play an intrinsic role in determining text classifier performance. Indeed, a large number of unique words and low overlap of vocabularies leads to higher uncertainty in text classifier predictions, where sub-word tokenisation proves insufficient. Despite their ability to model out-of-vocabulary words, we observe that models still get outdated. We quantify this by measuring the change of meanings of word and aspect representations over time in discrete language models trained over time. The good performance of contextual language models shows their strong potential, but in turn demonstrates that there is still room for improvement in furthering their temporal persistence. However, **contextual language models provide a solid alternative among the state-of-the-art solutions**.

## 7.2. Suggestions for the temporal evaluation of text classifiers

In our work we propose a novel evaluation framework which can effectively quantify the persistence of different text classifiers. Driven by the analysis of our results, we next provide a set of key suggestions to take into account for the design of experiments evaluating text classifiers with their temporal persistence as the key objective, as well as to enable furthering research in this direction:

- **Evaluating text classifiers across different time periods.** Where possible, evaluating models across test data pertaining to different time periods can help perform a more comprehensive assessment towards making models more generalisable and persistent over time. Where longitudinal datasets are not available, insights from our study, summarised in Section 7.1, can help design experiments with the temporal persistence as the objective.

- **Using complex longitudinal benchmarks** for evaluating the temporal persistence of models. As we observe in our research, there is substantial variation in the complexity of achieving temporal persistence. Indeed, persistence is relatively straightforward in a dataset exhibiting a more stable vocabulary, such as ABRR with Amazon book reviews, and much more complicated for datasets exhibiting more linguistic variations, as is the case with the social media datasets GESD and TESA. Hence, in order to make a fair assessment of models over time, it is ideal to include a range of datasets of different levels of complexity when it comes to language variation.

- **Studying temporally adaptive architectures** that can gradually learn with continually evolving datasets, including supervised models, but ideally focusing on unsupervised models, provided that labels are rarely available for large-scale, longitudinal datasets.

- **Making the most of existing pretrained language models.** Given the cost, complexity and often prohibitive use of computational resources to train new models (Treviso, Ji, Lee, van Aken, Cao, Ciosici, Hassid, Heafield, Hooker, Martins et al., 2022), it is also important to consider the efficiency of existing models in making them persistent over time. Improvement over temporal consistent would be ideal if it is achieved through computationally efficient resources that do not need expensive resources to train and that can be accessible to a wider range of users.

### 7.3. Limitations

Our study presents a comprehensive analysis of a wide range of combinations of language models and algorithms across three longitudinal datasets. However, in using this wide range of combinations, our objective has not been to achieve the best possible performance in each dataset, but instead to extensively assess temporal persistence through comparative experiments. In doing so, we have not conducted extensive hyperparameter tuning and one could expect that higher performance scores could be achieved by further tuning the models, but not necessarily improving temporal persistence. Likewise, in focusing on models tested across years, we assess model generalisability over time by exclusively leveraging knowledge acquired from the training data. To further leverage unlabelled data from subsequent years, one could consider further investigation into alignment methods (Alkhalifa et al., 2021).

While splitting our datasets into blocks of one year each and experimenting across these years, we present the aggregated results for sets of years which are the same number of years apart from each other, e.g. a gap of 1 year between training and test data aggregates results for a set of years that are one year apart from each other. Consequently, it is worth noting that experiments in larger temporal gaps have fewer experiments to aggregate and can be less stabilised (e.g. the 18-year gap in ABRR includes only experiments between 2000 and 2018, whereas the 1-year gap combines 2000-2001, 2001-2002, 2002-2003, etc.).

Last but not least, and largely constrained by dataset availability, our experiments focus all on binary classification experiments. We can expect that the findings from this study would largely extend to multiclass classification experiments too. However, additional experiments would be needed to further confirm this.

## 8. Conclusion

Despite the evidence of model performance deterioration over time due to changes in data, previous research did not delve into the factors impacting this deterioration. Through extensive exploration of and experimentation on three longitudinal temporal datasets, our comprehensive analysis provides insights into the role of five dimensions on temporal performance: language representations, classification algorithms, time, lexical features and context. Our study identifies the main challenges to focus on towards achieving temporally persistent classification models. While models generally struggle with persisting performance over time, we observe that RoBERTa shows a more generalisable performance across different datasets, improving over other models such as BERT. Despite these differences, we observe that different language models and algorithms consistently underperform on temporally distant test data, which is particularly impacted by the degree of variability of the dataset in question; while more closed-domain datasets such as book reviews show more stable performance over time, models struggle particularly with open-domain datasets exhibiting more prominent changes in language use. Our study provides important insights into model deterioration that help inform further research in the development of temporally-stable classification models.

While our study has focused on understanding the capacity and limitations of widely-used classification approaches, and drawing a set of best practices from this analysis, future research could further look into tackling the problem through domain adaptation and transfer learning. We also aim to study the computation complexity and effective design for hyperparameter tuning to support classifier persistence over time.

| DATASETS | | | | | |
|---|---|---|---|---|---|
| **GESD** | | **TESA** | | **ABRR** | |
| ***BERT-based aspects temporal cosine similarities variance*** | | | | | |
| **common aspects** | $\sigma^2$ | **common aspects** | $\sigma^2$ | **common aspects** | $\sigma^2$ |
| dowry death | $0,0065$ | most welcome | $0,0092$ | the bad piece | $0,0019$ |
| interesting report | $0,0064$ | I miss you | $0,0068$ | super book | $0,0018$ |
| amazing group of woman | $0,0014$ | the good | $0,0026$ | and entertaining book | $0,0005$ |
| male hatred | $0,0014$ | head hurt so bad | $0,0026$ | very disappointing book | $0,0004$ |
| **seasonal aspects** | $\sigma^2$ | **seasonal aspects** | $\sigma^2$ | **seasonal aspects** | $\sigma^2$ |
| husband ask why | $0,0110$ | love you | $0,0143$ | in timely manner | $0,0122$ |
| misandry strike | $0,0072$ | thank mummy | $0,0076$ | good seller list | $0,0093$ |
| I see they | $0,0011$ | I need food | $0,0018$ | protect a family | $0,0003$ |
| the legal protection | $0,0010$ | I love grill | $0,0018$ | they claim he be | $0,0003$ |
| **unique aspects** | $\sigma^2$ | **unique aspects** | $\sigma^2$ | **unique aspects** | $\sigma^2$ |
| gap well | $0,0074$ | @realDonaldTrump great | $0,0341$ | a prompt fashion | $0,013$ |
| be willing to make mistake | $0,0058$ | just stunned | $0,0097$ | in true geoff johns | $0,0106$ |
| fantastic group of college | $0,0011$ | just love that baseline | $0,0015$ | he use apparently british slang | $0,0003$ |
| the quiet recess | $0,0010$ | I wish they tape | $0,0015$ | fun and offer practical | $0,0003$ |
| ***Roberta-based aspects temporal cosine similarity variance*** | | | | | |
| **common aspects** | $\sigma^2$ | **common aspects** | $\sigma^2$ | **common aspects** | $\sigma^2$ |
| political empowerment | $0,0032$ | lovely lady | $0,0031$ | a complete rip | $0,0001$ |
| so inspiring | $0,0019$ | so handsome | $0,0026$ | m sorry to say that | $0,0001$ |
| amazing group of woman | $0,0002$ | miss they so | $0,0006$ | big waste of time | $0,0000$ |
| to strong woman | $0,0001$ | so so sad | $0,0005$ | mess of a book | $0,0000$ |
| **seasonal aspects** | $\sigma^2$ | **seasonal aspects** | $\sigma^2$ | **seasonal aspects** | $\sigma^2$ |
| never be more relevant | $0,0018$ | good beauty | $0,0042$ | significant character | $0,0007$ |
| racist people | $0,0013$ | be as bright | $0,0031$ | strong magnifying | $0,0004$ |
| sick of | $0,0002$ | really wanna get | $0,0003$ | I really appreciate the way | $0,0000$ |
| work on the basis | $0,0002$ | plan for this weekend | $0,0002$ | compliment the story | $0,0000$ |
| **unique aspects** | $\sigma^2$ | **unique aspects** | $\sigma^2$ | **unique aspects** | $\sigma^2$ |
| an epic fail | $0,0029$ | good nazi | $0,0036$ | due to damage | $0,0006$ |
| breach of the FA | $0,0023$ | the correct response | $0,0035$ | use editorial help | $0,0005$ |
| sensitised student on issue | $0,0001$ | I honestly wish | $0,0003$ | all about the bullying | $0,0000$ |
| to false rape victim | $0,0001$ | quiet about kmm | $0,0002$ | quite right with this book | $0,0000$ |

**Table A1**
Aspects represented using BERT and RoBERTA using high and low contextual variability score examples for each dynamic language use type.

# Appendix

## A. Aspects variance

Complementing the analysis in Section 6.5, Table A1 shows specific examples of aspects for each dataset, grouped by type. The table shows the two aspects with the highest and the two with the lowest similarity variations.

## Acknowledgment

This research utilised Queen Mary's Apocrita HPC facility, supported by QMUL Research-IT.

**Rabab Alkhalifa**, Ph.D. student at the Queen Mary University London. Her dissertation looks into the peculiarities of temporal data mining and the persistence of machine learning models performance over time. Rabab's study is funded by Imam Abdulrahman bin Faisal University, where she also works as a lecturer in the Computer Engineering department. Before starting her Ph.D., she graduated from the University of Bristol with a Master of Science in Advanced Computing - Machine Learning, Data Mining, and High-Performance Computing.

**Dr. Elena Kochkina**, Postdoctoral Researcher at the Queen Mary University of London and the Alan Turing Institute. Elena have completed a PhD in Computer Science at the Warwick Institute for the Science of Cities (WISC) CDT, funded by the Leverhulme Trust via the Bridges Programme. Her background is Applied Mathematics (BSc, MSc, Lobachevsky State University of Nizhny Novgorod) and Complexity Science (MSc, University of Warwick, Chalmers University). She has published in venues such as ACL, COLING, EACL and IP&M.

**Dr. Arkaitz Zubiaga**, Senior Lecturer at the Queen Mary University of London, where he leads the Social Data Science lab. His research interests revolve around linking online data with events in the real world, among others for tackling problematic issues on the Web and social media that can have a damaging effect on individuals or society at large, such as hate speech, misinformation, inequality, biases and other forms of online harm. He has published over 130 research papers (including 50+ journal articles), serves as academic editor for six journals and as SPC for a range of conferences in computational social science, natural language processing and artificial intelligence.